\documentclass[runningheads, anonymous=true]{llncs}
\usepackage{graphicx}
\usepackage{booktabs}
\usepackage{multirow}
\usepackage{array}
\usepackage{threeparttable}

\newcolumntype{x}[1]{>{\centering\arraybackslash\hspace{0pt}}m{#1}}
\newcolumntype{z}[1]{>{\centering\arraybackslash\vspace{0.3pt}}m{#1}}
\newcolumntype{y}[1]{>{\centering\arraybackslash}m{#1}}
\newcolumntype{q}[1]{>{\arraybackslash\vspace{0pt}}p{#1}}

\begin{document}
\title{Automated Code Extraction from Discussion Board Text Dataset}
%
%
\author{Sina Mahdipour Saravani\inst{1}\orcidID{0000-0003-4285-1439} \and
Sadaf Ghaffari\inst{1}\orcidID{0000-0002-2413-5829} \and
Yanye Luther\inst{1}\orcidID{0000-0002-0995-3356} \and
James Folkestad\inst{1}\orcidID{0000-0003-0301-8364} \and
Marcia Moraes\inst{1}\orcidID{0000-0002-9652-3011}
\email{\{sinamps, sadaf.ghaffari, yanye.luther, james.folkestad, marcia.moraes\}@colostate.edu}
}
\authorrunning{F. Author et al.}
%
\institute{Colorado State University, Fort Collins CO 80523, USA}
\authorrunning{F. Author et al.}
%

%
\maketitle              
\begin{abstract}
This study introduces and investigates the capabilities of three different text mining approaches, namely Latent Semantic Analysis, Latent Dirichlet Analysis, and Clustering Word Vectors, for automating code extraction from a relatively small discussion board dataset. We compare the outputs of each algorithm with a previous dataset that was manually coded by two human raters. The results show that even with a relatively small dataset, automated approaches can be an asset to course instructors by extracting some of the discussion codes, which can be used in Epistemic Network Analysis.

\end{abstract}
\section{Introduction}
Epistemic Network Analysis (ENA) has been used as an analysis technique in several different domains such as health care~\cite{zorgHo2021mapping}, educational games~\cite{bressler2021stem}, and online discussion~\cite{rolim2019network}. Recently, some researchers started to use ENA as a learning analytics visualization to support participatory Quantitative Ethnography (QE)~\cite{vega2021constructing} and instructor’s assessment of student’s participation in online discussions~\cite{moraes2021using}. In that work, researchers presented the use of ENA as a tool to visualize connections among the codes that were discussed in the online discussion boards to assist instructors in assessing those discussions. Two human annotators manually coded the text data.

Based on the study presented in~\cite{moraes2021using}, we submitted a project proposal to our university's teaching innovation grant to examine the use of ENA as a visualization tool. One feedback provided by the reviewers was that instructors would not have time to be involved in the coding process, even if that process used nCoder~\cite{ncoder}. The reviewers pointed out that they would like to have access to the visualization but not have the “burden” to build the codes. However, they would be willing to provide keywords that should be present in the codes. Considering that feedback, we decided to apply text mining and Natural Language Processing (NLP) algorithms to automate the extraction of codes from data and use the keywords that were provided by the instructors as guiding input to the algorithms. NLP has been applied to various applications~\cite{esmaeilzadeh2022efficient,zuo2022seeing} and has improved drastically with deep learning~\cite{sinamsthesis,saravani2021investigation,saravani2021automated}, motivating us to exploit it in this educational application. 

We developed three text mining systems based on Latent Semantic Analysis, Latent Dirichlet Analysis, and Clustering Word Vectors to examine their capabilities for this task and determine the one that produced the best results. We compared the codes that were previously manually coded and validated by human coders in~\cite{moraes2021using} with the outputs from the automated systems.

Previous work~\cite{bakharia2019equivalence} on comparing coding from human raters and automatic algorithms found that the topic modeling algorithm was only compared with human coders for broad topics and that additional domain knowledge is necessary to identify more fine-grained topics. Similarly, another work~\cite{cai2021using} examined how keywords from codes manually extracted were presented in topics generated from topic modeling. It found that top keywords of a single topic often contain words from multiple manual codes. Contrarily, words from manual codes appear as high-probability keywords in multiple topics. It is important to notice that both studies used large datasets to run their topic modelling algorithms. Our study is different in the sense that we have a relatively small dataset (section~\ref{sec:dataset}). We will consider the findings from both previous works and will discuss how those are present in our work.

In this paper, we present a set of potential solutions to automatically extract codes from discussion data. We also propose detailed modifications to previously incorporated methods as a way to overcome the limitation of working with a small dataset. This is a challenging problem since there is no accurate mapping between this task and the well-defined tasks in Natural Language Processing (NLP) and Information Retrieval. This results in a lack of solid prior studies on the topic. We demonstrate that even with a small dataset, automated approaches can be an asset to course instructors by extracting some of the discussion codes. Further, instructors can provide input to the algorithms to guide them towards better results.

\subsection{Research Questions}
Considering the particularity of our study regarding the use of a small dataset,  we aim to answer the following research questions:
\begin{itemize}\item RQ1: Are Latent Semantic Analysis, Latent Dirichlet Analysis, and Clustering Word Vectors algorithms able to extract the codes previously coded by~\cite{moraes2021using}?\item RQ2: What are the limitations of each algorithm? 
\end{itemize}

\subsection{Dataset}\label{sec:dataset}
Our dataset consisted of online discussion data from seven semesters (Fall 2017, Fall 2018, Fall 2019, Spring 2020, Fall 2020, Spring 2021, Fall 2021) from an online class for organizational leaders as part of a Masters of Education program at a Research 1 land-grant university. The dataset has 2648 postings. This dataset was provided to us by Moraes and colleagues~\cite{moraes2021using}. The codebook established after meeting with Moraes and colleagues consists of the “a priori” codes presented in Table 1.

\begin{table}[!tp]
    \caption{Codebook}
    \centering
    \setlength{\tabcolsep}{5pt}
    \begin{tabular}{m{0.2\linewidth}m{0.55\linewidth}x{0.1\linewidth}} 
    \toprule
        \textbf{Code Name} & \textbf{Definition} & \textbf{Kappa}\\
        \cmidrule{1-3}
        Retrieval practice, Spacing out practice, Interleaving
        &
        Retrieval practice is the act of recalling facts or concepts or events from memory. The use of retrieval practice as a learning tool is known also as testing effect or retrieval-practice effect. Spacing out practice allows people to a little forgetting that helps their process of consolidation (in which memory traces are strengthened, given meaning, and connected to prior knowledge). Interleaving the practice of two or more concepts or skills help develop the ability to discriminate later between different kinds of problems and select the better solution. 
        &
        0.85\\
        \cmidrule{1-3}
        Elaboration
        &
        Elaboration is the process of giving new material meaning by expressing it in your own words and making connections with your prior knowledge. The more connections you create, the easy will be to remember it in the future. 
        & 0.79\\
        \cmidrule{1-3}
        Illusion of mastery
        &
        Researches have pointed out that students usually have a misunderstanding about how learning occurs and engage with learning strategies that are not beneficial for their long-term retention, such as rereading the material several times and cramming before exams. When we got familiar with some content due to fluency in reading it, we create an illusion of mastering that content.
        &
        0.89\\
        \cmidrule{1-3}
        Effortful learning
        &
        Learning is deeper and more durable when it is effortful, meaning that efforts, short-terms impediments (desirable difficulties), learning from mistakes, and trying to solve some problem before knowing the correct answer makes for stronger learning.
        &
        0.85\\
        \cmidrule{1-3}
        Get beyond learning styles
        &
        There is no empirical evidence of the validity of learning styles theory in education. Researchers found that when instructional style matches the nature of the content, all learners learn better, regardless of their learning styles.
        &
        0.86\\
    \bottomrule
    \end{tabular}
    \label{table:dataset}
\end{table}

\section{Related Work}

Previous works such as~\cite{cai2021using} in the context of topic modelling mostly deal with using nCoder. Although nCoder is a popular learning analytic platform used to develop a coding scheme, it is not fully automated. In other words, it requires human in the loop to read through the text and validate if the coding is appropriate. nCoder+ ~\cite{cai2019ncoder+} is another work in the direction of nCoder. This work specifically discusses an improvement to nCoder through semantic component addition to the nCoder process. This paper contributes to finding false negatives in the current nCoder version. Furthermore, as mentioned by authors of nCoder+, the idea is just a prototype and is not a public tool yet. A very recent work by Cai et. al~\cite{cai2021using} centers around nCoder as well. They investigate how close human created codewords are to topic models in large text documents. They also discuss whether the top keywords in topics match human codewords. There are distinct aspects between our work and others. First, we mainly take advantage of NLP approaches and compare three different unsupervised learning techniques to automate the topic extraction in our relatively small text documents. Second, we utilize coherence analysis as a way to determine the optimal number of topics in our text discussion data. Interestingly, in the context of our data, the established number of topics is same as our acquired ground truth information from course instructors. In Section~\ref{sec:methods}, we provide the details of our techniques for topic extraction.




\section{Methods}\label{sec:methods}

For mining topics from our textual data, we begin our analysis with applying Latent Semantic Analysis (LSA)~\cite{landauer1998introduction} to our discussion data. We then present two methods for automatic code extraction from text data. The first method follows the efforts of previous researchers and is based on Latent Dirichlet Allocation (LDA) topic modeling~\cite{blei2003latent}. The second method uses K-means~\cite{macqueen1967some} to cluster word embeddings. Both of these methods provide a way for the expert user to guide the automated learning process. Further details are provided in the following sub-sections.

\subsection{Shared Preprocessing for All Algorithms}
Preprocessing is highly important in NLP tasks that are based on the bag of words modeling of the text where the order of words in the sentences is ignored and a sentence is considered a bag of its words. Since our target task entails the retrieval of codes -- in contrast with classification tasks like \cite{saravani2021automated} -- and the best outcome is to retrieve each of them in a uniform written form, such preprocessing steps are critical for better performance. Stop words, for example, do not carry any information that is useful for our task. Named-entity categories such as the name of authors and dates are also not relevant to the codes. We performed the following preprocessing steps that improved our experimental results. All of these steps, except the last one, are also applied in the word embedding method. The term document refers to a post in our online discussion board dataset.
\begin{itemize}
    \item \textit{Tokenization}. We use the \textit{simple\_preprocess} from GENSIM \cite{rehurek_lrec} to tokenize the text using a regular expression from Python regex package. 
    \item \textit{Lowercasing}. We lowercase all words for uniformity and normalization of various writing styles and arbitrary capitalization.
    \item \textit{Stop word removal}. Stop words do not contribute to codes and their removal would allow the algorithms to focus on useful words.
    \item \textit{Applying minimum word length}. We assume that words with a length less than 3 characters in the English language do not contain useful information.
    \item \textit{Irrelevant text removal}. We remove non-breaking space (NBSP), roman numerals, numbers, and URLs using regular expressions since these components do not contribute to the codes.
    \item \textit{Named-entity removal}. Names of the students, authors, or books are not relevant to codes. After preliminary experiments, we concluded that they confuse the algorithms. We use SpaCy~\cite{matthew_honnibal_2019_3358113} to remove all tokens that are recognized as the name of a person or a work of art.
    \item \textit{In-document frequency filtering}. Words that very rarely appear or are very frequent are not likely to be codes. We removed words that occur only in one single document. We also removed words that occur in more than 10\% of the documents.
    \item \textit{Generating bigrams and trigrams}. Since we are interested in code terms that contain more than one word -- e.g. the code term ``interleaved practice'' -- we extend the list of tokens in each document with its bigrams and trigrams: consecutive two and three words \cite{esmaeilzadeh2022building}.
\end{itemize}

In our second proposed method, clustering word vectors: fastText + K-means, we do not consider bigrams and trigrams due to the lack of a natural operator to represent two or three single word vectors as a single vector of the same dimension. We leave further explorations of this path to future work.

\subsection{Code Extraction Algorithms}\label{sec:algorithms}
One potential solution to the problem of code extraction, following the related work~\cite{cai2021using}, involves using topic modeling algorithms. Another potential solution involves using word embedding vectors and clustering algorithms. We present a brief introduction to these algorithms and explain how we use them to discover the different topics of discussion in the class. Further, we provide qualitative comparisons between them for our objective.

\subsubsection{Latent Semantic Analysis (LSA)} is a technique discovering statistical co-occurrences of words which appear together and provide further insights to the topic of words and documents. In this approach, a term-document matrix is formed from documents. The rows, in this matrix, are individual words and columns are documents. Specifically, the entries in a term-document matrix contain the frequency which a term occurs in a document. Singular Value Decomposition (SVD) is applied to the Term-Document matrix. As a result of applying SVD, we get the best k-dimensional approximation to the term-document matrix. The similarities among the entities in the lower dimension space are computed. 

\subsubsection{Latent Dirichlet Analysis (LDA)} is a generative probabilistic model in the context of NLP which represents documents over latent topics. These topics are each a distribution over words.
The main assumption in topic modeling is that documents discuss the same topic if they use the same or similar set of words. Topics in LDA act as a hidden intermediate level between documents and words. This provides a path to dimensionality reduction, where a large matrix of association between all documents and words is broken into two matrices: one between documents and latent topics, and one between the latent topics and words~\cite{lda2018intuitive,lda2021topic}.
Being built upon the probabilistic latent semantic indexing (pLSI ~\cite{hofmann1999probabilistic}, LDA overcomes some of the shortcomings of pLSI. The pLSI has no natural way to assign a topic probability to an unseen document since it only learns a topic-document distribution for documents it has seen in its training. This is due to the use of a distribution that is indexed by training documents. Growth of pLSI's number of parameters with the number of documents also makes it prone to overfitting. 
Both of these limitations are resolved in LDA by using a parameterized hidden random variable as a topic-document distribution which is not explicitly linked to training documents.

LDA and the task of topic modeling for automated code extraction benefits from the separation of documents. The separation of documents is actually inherent in our target task. We intend to i) know what codes are covered in each discussion post, i.e. in each document, and ii) extract all codes that have appeared in all discussions. Hence, the separation of documents provides useful information for our objective.

After preprocessing, we create a LDA topic model over our aforementioned dataset (Sec.~\ref{sec:dataset}). We experiment with three variations: \textit{without bigrams and trigrams}, \textit{with bigrams and trigrams}, and \textit{with bigrams and trigrams where prior topic-word distribution is modified}. The first two variations are self-explanatory. The third variation introduces the notion of a human user providing some codes or keywords in an input file to the program that are likely to be discussed in the course. We assume a constant topic for all keywords that are provided in a single line in the input file. For all keywords in a line, we assign a higher probability of belonging to that arbitrary constant topic compared to all other words in the dataset. We currently use a hyperparameter value (\textit{keywords\_total\_probability}) to divide over the number of keywords and assign it to each of them. Thus, the probability for each keyword-topic pair is:
\begin{equation}
    p(k_t,t) = \frac{\textit{keywords\_total\_probability}}{n_{k_t}}.
\end{equation}

For the rest of the words, the probability for that topic is the following uniform value,
\begin{equation}
    p(w,t) = \frac{(1-\textit{keywords\_total\_probability})}{(n_w - n_{k_t})},
\end{equation}
where $n_w$ is the number of words in the whole dataset and $n_{k_t}$ is the number keywords that user has provided for the arbitrary topic t. $k_t$ is a provided keyword for topic $t$.

In case there are topics that the user has not provided any keywords for, the word-topic probability is
\begin{equation}
    p(w,t) = \frac{1}{n_w}.
\end{equation}

Although our experiments had a limited volume of data, this mechanism does not provide a significant improvement but provides a natural way to optionally use human expert knowledge in the code extraction process using LDA. We implement this mechanism using the \textit{eta} parameter of GENSIM's LDA implementation.

\subsubsection{Clustering Word Vectors: fastText + K-means} is another method which we present here as a potential solution to the code extraction problem. This approach is based on the idea of clustering algorithms. Although the code retrieval process is not evident, we propose a heuristic for that. The first requirement for clustering is representing the words in vectors. However, bag of words modelling is not sufficient here, as the clustering algorithm, being an unsupervised algorithm, only captures the already-present distinguishing characteristics of data rather than learning new features. Hence, the word vectors need to carry useful information for the latter use of the clustering algorithm. Exploiting the syntactic and semantic information contained in pre-trained word embedding vectors, which have successfully improved the accuracy on many NLP tasks ~\cite{mikolov2013efficient,bojanowski2016enriching,pennington2014glove,devlin2019bert}, is a natural choice. To further handle out-of-vocabulary words that have not been observed in the pre-training process, we use fastText~\cite{bojanowski2016enriching} word embeddings, which is a character-level word representation model. This ensures that even uncommon words that are discussed in a specific expert course would have appropriate vector embeddings. After preprocessing, we convert all of the words in the dataset to their embedding vectors. Unfortunately, in this step, we lose the information about which documents each word appeared in, which is a limitation of this approach. Next, we use the K-means clustering algorithm~\cite{macqueen1967some} to create groups of semantically and syntactically relevant words.

Word embedding models are pre-trained on huge amounts of digital text to learn the co-occurrence statistics of words. This pre-training objective results in a projection of syntactically and semantically relevant words to a close proximity in the resulting vector space. Furthermore, these word vectors have an interesting behaviour against addition and subtraction operators where they prove capable of learning some relationships among words. For example, the arithmetic operation $\textit{Paris - France + Italy}$ equals $\textit{Rome}$ \cite{mikolov2013efficient}.

For the retrieval of words that represent each cluster, we utilize cluster centroids. We find the five closest word vectors to the center of each cluster and retrieve their words as the extracted codes. The closeness measure is the cosine similarity function. While this mechanism extracts interesting words, it has some limitations. The distance from cluster centroids does not necessarily imply irrelevance. Some code words may be closely related to two topics and far from the center of both associated cluster; hence, they would not be retrieved in either of the clusters. On the other hand, there is no reason to believe that the code words would appear in the center of the clusters.

For an analogous use of human expert input, the program accepts the same aforementioned keywords file. Here the input keywords, if present, are used to initialize the K-means cluster centroids. First, all words in a single line of the keywords file, which are assumed to represent one code, are converted to their respective fastText word embeddings. Then they are averaged and used as the initial center for one arbitrary cluster. During the training process of K-means, they are updated to fit the dataset. Generally, K-means algorithm without specific initialization is run multiple times with random initialization to find the best final convergence. However, since we specifically initialize all or some of the centroids using the keywords file, this variation of our implementation would execute K-means only once. In this case, it is reasonable to expect the code words to appear at the center of the clusters. This mitigates the aforementioned concern about the centers being ill-defined in terms of representing the actual codes, and further provides more computational efficiency by avoiding redundancy~\cite{saravani2021investigation,sinamsthesis}.

The most important limitation of this approach is that the information about document boundaries and what words are contained is what documents are lost. While benefiting from the knowledge that is transferred from the pre-training of the word embedding model, this method treats the whole dataset merely as a dictionary of words.

\section{Experimental Results}
In order to answer the research questions we posed, we conducted experiments with the three aforementioned algorithms. To extract the topics from our discussion data, we were interested to find the optimal number of topics for our approach. Therefore, we conducted a topic coherence analysis~\cite{roder2015exploring} to achieve our aim. Given that text mining algorithms are in the category of unsupervised learning approaches, coherence analysis is considered an important measure as it gives structure to inherently unstructured textual data. Furthermore, it assesses the topics' quality and the coherence of words within each topic or cluster. The results in Table~\ref{tab:table1} indicate the optimal number of clusters is five. Therefore, we present ten words from each of the five topics for respective algorithms. The extracted topics from the Latent Semantic Analysis, Latent Dirichlet Allocation, and Clustering Word Vectors algorithms are demonstrated in Tables~\ref{tab:results-lsa}, ~\ref{tab:results-lda}, and~\ref{tab:results-w2v} respectively.

\begin{table}[h!]
  \begin{center}
    \caption{Coherence Score for Various Number of Topics.}
    \label{tab:table1}
    \begin{tabular}{x{2cm}x{2cm}}
      \toprule
      \textbf{No. Clusters} & \textbf{Coherence Score} \\
      \cmidrule(lr){1-2}
      2 & 0.2851 \\
      3 & 0.2915 \\
      4 & 0.2944 \\
      5 & 0.5017 \\ 
      6 & 0.3776 \\
      7 & 0.4071 \\
      8 & 0.4067 \\
      9 & 0.3427 \\
      10 & 0.3773 \\
      \bottomrule
    \end{tabular}
  \end{center}
\end{table}


\begin{table}[t] 
\caption{Extracted Topics from Latent Semantic Analysis}\label{tab:results-lsa}
\setlength{\tabcolsep}{16pt}
\centering
\begin{tabular}{ccccc} 
 \toprule
 Topic 0 & Topic 1 & Topic 2 & Topic 3 & Topic 4 \\ [0.8ex] 
 \cmidrule(lr){1-5}
 learn & practice & practice & memory & memory \\ 
 practice & learn & forget & forget & difficulty \\
 author & mass & mass & train & desire \\
 think & author & lecture & difficulty & train \\
 memory & time & inform & practice & example \\
 time & memory & think & desire & story \\
 wait & retrieve & learn & example & mindset \\
 inform & forget & understand & mass & deliberation \\
 knowledge & like & surgeon & learner & growth \\
 skill & train & solution & help & plf\textsuperscript{*} \\ [1ex]
 \bottomrule
\end{tabular}
\begin{tablenotes}
      \footnotesize
      \item $^*$Stands for Parachute Landing Fall.
    \end{tablenotes}
\end{table}

\begin{table} [hbt!]
\caption{Extracted Topics from Latent Dirichlet Allocation}\label{tab:results-lda}
\setlength{\tabcolsep}{3pt}
\centering
\small
\begin{tabular}{ccccc} 
 \toprule
 Topic 0 & Topic 1 & Topic 2 & Topic 3 & Topic 4 \\ [0.8ex] 
 \cmidrule(lr){1-5}
 lecture & desire & dylexia & confidence & mass \\ 
 solution & desire\_difficuty & learn\_style & feedback & mass\_practice \\
 classroom & plf & individual & calibration & interleaving\_practice \\
 surgeon & resonate & learn\_differ & confidence\_memory & space\_retrieval \\
 acquire & parachute & disable & accuracy & tend \\
 instruct & fall & intelligent & peer & day \\
 learn\_learn & land & prefer & answer & long\_term \\
 impact & jump & support & event & week \\
 demand & parachute\_land & dyslex & state & myth \\
 lecture\_classroom & land\_fall & focus & calibration\_learn & practice\_space \\ [1ex]
 \bottomrule
\end{tabular}
\end{table}

\begin{table} [hbt!]
\caption{Extracted Topics from fastText + K-means}\label{tab:results-w2v}
\setlength{\tabcolsep}{15pt}
\centering
\small
\begin{tabular}{ccccc} 
 \toprule
 Topic 0 & Topic 1 & Topic 2 & Topic 3 & Topic 4 \\ [0.8ex] 
 \cmidrule(lr){1-5}
 aspects & teacher & merely & pull & just \\ 
 concepts & girl & simply & off & even \\
 strategies & college & seemingly & down & know \\
 perception & american & consequently & sticking-out & think \\
 knowledge & school & ostensibly & underneath & going \\
 understanding & juniour & essentially & divits & get \\
 implications & teacher & evidently & loose & thought \\
 analysis & baylee & being & stick & really \\
 approach & mom & therefore & rope & come \\
 methodology & student & rather & sideways & but \\ [1ex]
 \bottomrule
\end{tabular}
\end{table}

To answer RQ1 and RQ2, we presented these results to Moraes and colleagues. Through qualitative analysis they determined that the outputs from LSA and LDA could extract some of the codes previously coded, but Clustering Word Vectors did not capture any code. They mentioned that the best code words are captured by the presented LDA algorithm where 4 of the 5 codes from the manual annotation have been extracted. In Table~\ref{tab:results-lda}, \textbf{Topic 1} code words were related to \emph{Effortful learning} code, \textbf{Topic 2} code words were related to \emph{Get beyond learning styles}, \textbf{Topic 3} code words were related to \emph{Illusion of mastery}, and \textbf{Topic 4} code words were related to \emph{Retrieval practice}, \emph{Spacing out practice}, and \emph{Interleaving}. Only \textbf{Topic 0} did not represent any one of the codes. It is important to note that the \emph{Elaboration} code, not present in any topic, had the lowest kappa between the two human coders (Table 1), meaning that even between human raters it wasn't an easy code to agree upon. Our results corroborate what was found in~\cite{bakharia2019equivalence}, as the topic modeling algorithm could only find broad topics. Assess whether that is enough, in terms of accurately predicting codes in the remaining of our dataset, is something that still needs to be examined.

Our LSA results present the same limitations found by~\cite{cai2021using} work. As we can observe in Table~\ref{tab:results-lsa}, extracted code words of a single topic often contain words from multiple manual codes and  words from manual codes appear as keywords in multiple topics.

Analyzing the extracted topics from the fastText + K-means algorithm, \textbf{Topic 2}, \textbf{Topic 3}, and \textbf{Topic 4} in Table~\ref{tab:results-w2v} are not actual codes. \textbf{Topic 2} seems to contain only adverbs. As we expected and discussed this in Section~\ref{sec:algorithms}, the pre-training process of the word embedding models (fastText) causes the vector space to have syntactically similar words positioned in a close proximity -- the semantic closeness is not the only positioning objective of the pre-training process. As a result, the clustering algorithm (K-means) groups such words in a cluster. This is undesirable for the purpose of extracting codes. We discuss a potential solution to this problem in Section~\ref{sec:conclusion}.

\section{Conclusion and Future Directions}\label{sec:conclusion}
The goal of this study is to investigate the capabilities of Latent Semantic Analysis, Latent Dirichlet Analysis, and Clustering Word Vectors to automatically extract codes from a relatively small discussion dataset. As previously stated, this is a challenging problem due to a lack of solid prior studies on the topic. 

As observed in the experiments, LDA was the best approach and further examinations will be done in order to assess the prediction capabilities for the remaining dataset. The use of non-contextualized word embeddings, such as fastText (the third presented method (clustering word vectors)) results in some clusters containing only syntactically relevant words -- \textbf{Topic 2} in Table~\ref{tab:results-w2v}. However, syntactic similarity of words may not be necessarily useful in our target task of extracting codes because of the pre-training objective of the word embedding model. One possible solution is to use recent contextualized word embeddings such as BERT \cite{devlin2019bert} as a substitute. Their word representation vectors are much richer in linguistic information and have been pre-trained to attend to important tokens in a text~\cite{vaswani2017attention}. Such models also open up opportunities to keep the word-document information, as they are designed to use the context to generate the word vectors. However, due to their dynamic nature, the retrieval process is not trivial. 

Another possible solution is exploiting the LDA2Vec model for code extraction. As mentioned in Section~\ref{sec:algorithms}, both the topic modeling and the clustering word embeddings have their advantages and disadvantages. To gain benefits from both, we intend to conduct experiments with LDA2Vec. Many advances in NLP focus on improving machines in processing text, while LDA2Vec aims to automatically extract information that is useful to from a large volume of text~\cite{moody2016mixing}. The LDA2Vec model is a modified version of the skip-gram word2vec model, where in addition to the pivot word vector a document vector is also exploited to predict the context, a modified architecture and pre-training objective. This makes LDA2Vec capable of capturing document-related information as well as word information. Since word vectors, document vectors, and topics are learned at the same time, they would be linked to each other and contain inherent information about the others.

Despite the limitations presented in this study which corroborates the findings of previous works~\cite{bakharia2019equivalence,cai2021using}, we demonstrate that even with a relatively small dataset, automated approaches can be an asset to course instructors. Our intention is to incentivize further discussions on how text mining algorithms can be used to extract codes, even from relatively small datasets.






%
%
%
\bibliographystyle{splncs04}
\bibliography{references}
\end{document}